\documentclass[runningheads]{llncs}
\usepackage{graphicx}
\usepackage{amsmath, amssymb}
\usepackage{float}
\usepackage{subcaption} 
\usepackage{booktabs}
\usepackage{tabularx}
\usepackage[hidelinks]{hyperref}

\begin{document}
\flushbottom

\title{CGMap: A Geospatially Aware Deep Learning Framework for Crop Gap Mapping Using UAV}
\titlerunning{Sugarcane Emergence and Gap Mapping}
\authorrunning{K. Sharma} 
\author{
Karan Sharma\orcidID{0009-0004-5371-7280}\inst{} \and
Rajiv Ranjan\orcidID{0009-0008-3248-7374}\inst{} \and
Dinesh Kumar\orcidID{0009-0000-4952-0742}\inst{} \and
Shashank Tamaskar\orcidID{0000-0002-9497-2817}\inst{}
}

\institute{
Robotics and Autonomous Systems, Plaksha University, Mohali, India \\
\email{karan.sharma@plaksha.edu.in, rajiv.ranjan@plaksha.edu.in, 
dinesh.kumar1@plaksha.edu.in,
shashank.tamaskar@plaksha.edu.in}
}

\maketitle

\begin{abstract}
Crop germination is primarily monitored by visual inspection and manual counting, which are prone to errors despite their crucial role in determining the eventual yield potential.
This paper highlights a deep learning based pipeline which uses object detection methods and drone imagery to assess and provide a precise count of sugarcane germination in fields. The approach uses a pre-trained AI model to find germinated plant sapling and identify gaps, also known as ``bald spots'', which restrict field productivity. The techniques used here are based on the YOLOV8 architecture, which was trained on a carefully selected dataset of UAV photos taken in Western Uttar Pradesh, India. Here, we bring upon a novel orientation-normalization technique that uses minimum Spanning Trees (MST) to account for variations in planting geometry, allowing for dependable row and column extraction across a variety of field layouts.
By converting detected saplings into spatial point clouds, emergence gaps can be inferred from the anticipated spacing between plants. A geospatial germination map exported in Well-Known Text (WKT) format is the end result, and it can be easily incorporated into GIS platforms used by sugar mills and agronomists to direct transplant initiatives. Timely interventions based on the insights can significantly increase yield, resulting in higher profits. Hence, support proper allocation of resources, avoid waste, and enhance long-term sustainability.


\end{abstract}

\section{Introduction}

Remote sensing has emerged as a transformative tool in modern agriculture, enabling large-scale, continuous, and non-destructive crop monitoring \cite{10700753,ranjan2024stubble}. The crop of interest in this paper, Sugarcane (\textit{Saccharum officinarum}) \cite{artschwager1958sugarcane} is a critical crop for the global agricultural economy, serving as the main source of sugar, ethanol, and related by-products. India, among other producing nations, is the second largest sugarcane grower in the world, accounting for more than 5000 metric tons (LMT), following Brazil \cite{PIBreportonsugarcane,10640589}. India has now grown to be the world's largest sugar producer and the second largest exporter, providing the livelihoods of over 50 million farmers and with almost 10\% of the rural population in its cultivation, processing and supply chain \cite{solomon2022indian}.

Despite its scale and strategic importance, sugarcane production in India faces persistent challenges in achieving a uniform cropping pattern, especially during the early germination phase \cite{nalawade2018sugarcane,singh2019factors,smit2010characterising}.  Uniform germination is a key determinant of eventual and final yield, as localized zones of missing or failed germination, colloquially known as "bald spots" or "skips" can lead to uneven density, suboptimal resource utilization, and significant yield losses. Bald spots can be attributed to a complex interplay of environmental, mechanical, and biological factors, making it necessary to detect them early and find measures to eliminate them

Collectively, these factors underscore the importance of the timely detection and spatial diagnosis of bald spots to allow corrective actions, such as transplantation, spot fertilization, or pest control strategies that are critical to maximize crop stand and overall field productivity. 
 If not identified and replanted early, it will contribute to yield reductions of up to 10--15 tons per hectare, a figure corroborated by field reports from farmers who estimate 20--30\% productivity losses due to undetected bald patches during the critical early weeks of growth \cite{yadav2024navigating,rahman2019trends,arun2022sugarcane,10640589}. 

Traditionally, the germination of sugarcane saplings assessment and bald spot identification is carried out by manual field inspections that are labor intensive, time-consuming, and infeasible on a large scale.The emergence of precision agriculture, particularly with unmanned aerial vehicles (UAVs) and drone based imaging has provided a great alternative \cite{10.1007/978-3-031-88217-3_30,10984153}. UAVs equipped with high-resolution cameras help rapidly collect data in large areas in single flights, enabling near-real-time crop monitoring. The main challenge lies in converting raw high resolution imagery into agronomic insights, these might require techniques like AI models, computer vision and spatial analysis \cite{10.1145/3674829.3675079}. 

While the use of this technology has seen an emergence in Indian precision agriculture, it remains limited to detection of pests, disease etc. There is a dearth of studies that perform automated detection of germination gaps. This one of a kind study attemps to address this issue by highlighting a scalable, automated, and adaptive pipeline that uses an
object detection model for the detection of sugarcane saplings, OpenCV \cite{bradski2008learning} for row and column detection, and Rasterio \cite{rasterio} for spatial analysis. This helps us to detect individual sugarcane saplings from drone images and find bald spots with precision. A novel orientation correction method using MSTs \cite{pettie2002optimal} is also implemented to standardize field geometry and extract meaningful row-column structure data even in irregular or an off-axis farms. The germination maps are then exported in Geographical Information System (GIS)-compatible formats, providing a quick and reliable integration with farm management systems and decision-making tools for agronomy.

By automating this bald spot detection during the early growth window, this pipeline empowers the farm managers and the sugar mills to initiate timely transplantation on the farms, improving field uniformity and yield potential substantially. While this study is focused on sugarcane in north central India and uses YOLOv8 as the object detection model, the pipeline can be incorporated with any other object detection AI models (e.g., Faster R-CNN, EfficientDet), and is also transferable to other row crops. This illustrates the robustness of the pipeline and makes a contribution to the broader adoption of data-driven and sustainable agriculture.

\section{Materials and Methods}

\subsection{Study Area, UAV Platform and Image Acquisition}

Our study focuses on Sugarcane fields in Asmoli, Uttar Pradesh, India, located between 23°52'–30°28' N latitudes and 77°05'–84°38' E longitudes in fertile alluvial plains of the Indo-Gangetic floodplain of north central India, with elevations ranging from approximately 200 to 300m above sea level. The region’s soils are predominantly sandy loam to clay loam, known for their high agricultural potential, although they feature localized sandy and loess deposits \cite{Charankumar_Dhyani_Vivek_Shahi_Kumar_Singh_2022}. While generally fertile, parts of the area face salinity and alkalinity issues, primarily due to over-irrigation \cite{Dhanda2022}.
\sloppy
Despite these climatic and edaphic challenges, the combination of deep alluvial soils, effective irrigation infrastructure, and ample monsoonal rainfall supports the intensive cultivation of sugarcane, wheat, and rice. Sugarcane, in particular, thrives under these conditions. However, uniform crop emergence continues to be hindered by inconsistencies in summer sowing conditions, underscoring the need for timely agronomic interventions.

Phantom 4 Multispectral drone, developed by DJI was used to capture the UAV imagery for this study. As highlighted and optimized in previous literature, image acquisition for this study was conducted at flight altitudes of 37.8 and 75.6 meters \cite{mesas2015assessing}, which yielded a ground sampling distance (GSD) of approximately 2 cm/pixel and 4 cm/pixel. In order to ensure sufficient coverage and photogrammetric accuracy, a front overlap of 80\% and a side overlap of 75\% were maintained during all flights.

\subsection{Orthomosaic Generation and Image Preprocessing}

Pix4D Fields , a commercial photogrammetry software was used to process the raw aerial images acquired from UAV flights and to generate high-resolution, orthorectified mosaics (orthomosaics) for each surveyed field. Spatial geometry and radiometric fidelity were preserved in these orthomosaics for providing georeferenced 2D maps suitable for pixel-accurate object detection.

To prepare the data for model training and inference, the following preprocessing steps were applied:

Each orthomosaic was divided into smaller image tiles of size $300 \times 300$ pixels using a sliding window approach with 10\% overlap to preserve boundary information. Tiles with poor visibility due to shadows, flooding, or motion blur were manually filtered out. Finally, all selected tiles were converted to standard RGB format and resized to $640 \times 640$ pixels to match YOLOv8 input dimensions.

The resulting image set formed the input dataset for object detection and annotation.
\begin{figure}[H]
    \centering
    \includegraphics[width=\linewidth]{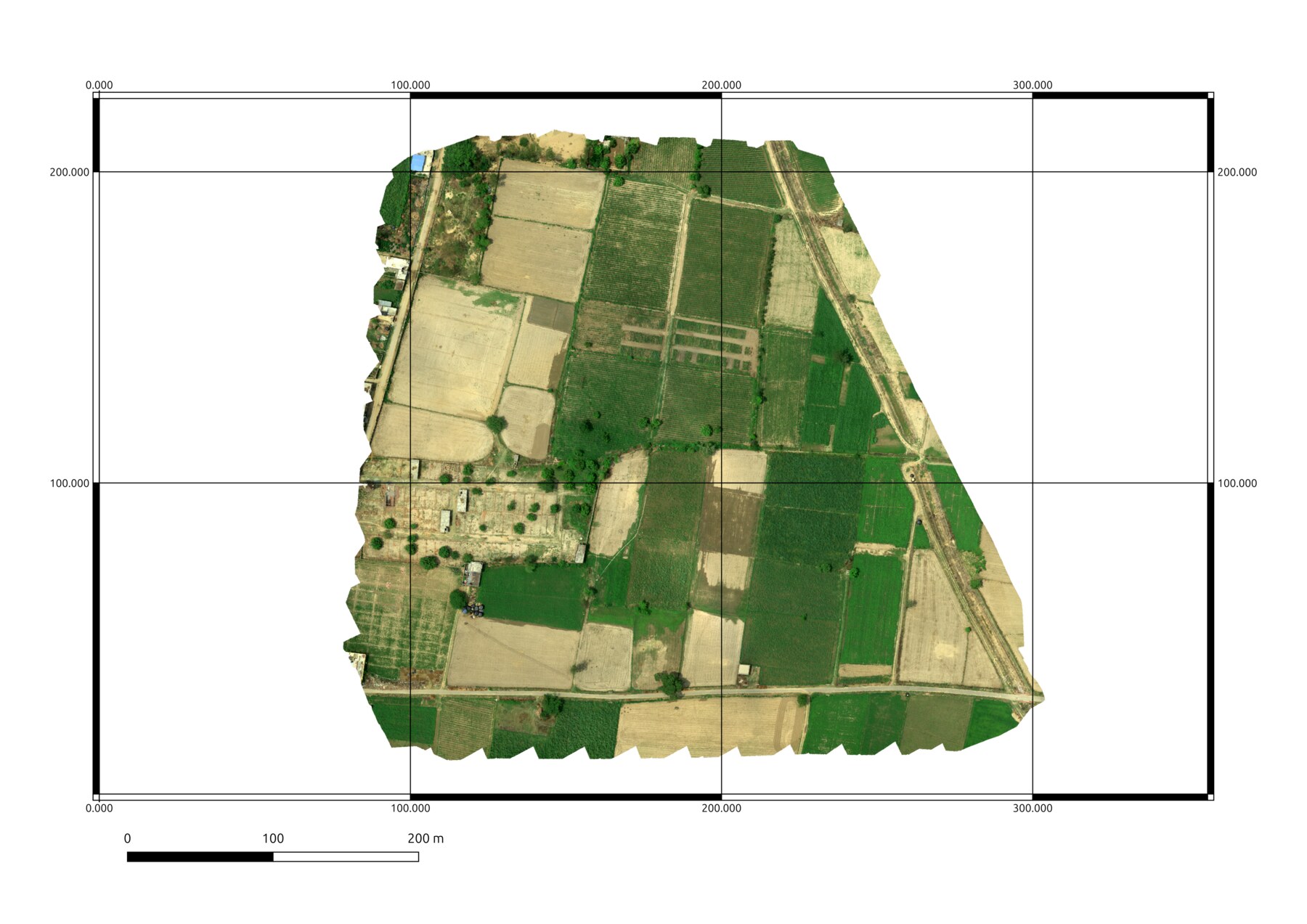}
    \caption{Georeferenced stitched orthomosaic obtained from UAV.}
    \label{fig:stitched orthomosaic}
\end{figure}
Despite the advancements in automated crop monitoring through UAV and AI based methods, majority of the studies in nations such as Brazil \cite{wachholz2017mapping}, following a large scale, agricultural system are developed in continuous uniform farmlands. These environments simplify image segmentation and model generalisation as they allow for consistent, planting geometry, predictable row, orientation, and homogeneous soil and canopy backgrounds. In contrast to this, Indian farmlands are fragmented, typically smaller and highly heterogeneous with irregular planting patterns with variable soil intercropping and inconsistent management practices \cite{farm_size}. These variabilities bring about a significant challenge for AI based models trained on data from large uniform fields as a shift in spacial and spectral conditions often deteriorates their performance. These irregularities may be attributed to the limited mechanisation in Indian agriculture due to the small average farm sizes and a restricted adoption of agricultural technologies. As clearly illustrated in figure \ref{fig:stitched orthomosaic}, Asmoli, the area under study growing sugarcane, has large portions of land occupied by different crops, such as rice, potato, and are also left unsown making it hard to detect sugarcane growth with reliability solely via spacial or spectral cues. Hence, models from the western world might not be so neatly transferable in an Indian context without adaptation or retraining. There is a need for developing robust, adaptable and region, specific models that can generalise across India’s diverse agricultural landscape.

\subsection{Dataset Annotation and Model Training}

Approximately 600 image tiles were annotated manually using the open-source web tool \texttt{Makesense.AI}. Each visible sugarcane sapling was bounded with a rectangular box and labeled as class \texttt{sugarcane}. The annotations were exported in {YOLO format}.

A model was required that could detect densely spaced plant instances in varying light and background conditions with high accuracy, computational efficiency, and could be easily deployed. These criteria were effectively fulfilled by YOLOv8 making it a good fit for large-scale agricultural field analysis. In addition to this, it is open source, benefits from continuous community driven improvements, has broad accessibility and a transparent development. This extensive support ecosystem and the availability of pre-trained weights, reinforce its reliability and adaptability across diverse applications.
Even though more advanced models may be used as substitutes, this proposed pipeline has been developed to remain model, diagnostic, resilient, while ensuring consistent performance, even when paired with detectors of relatively modest accuracy. this adaptability has been demonstrated by the use of YOLOv8 a general purpose detector that yielded that satisfactory and reliable results in the agricultural context. YOLOv8’s single stage, real time, detection framework, anchor, free design and efficient feature extraction together, allow for accurate identification of plant instances in complex and changing field environments.

Model convergence was monitored using validation mean Average Precision (mAP). Post-training evaluation indicated reasonable detection performance, suitable for field-scale deployment.

\begin{table}[H]
\centering
\begin{tabularx}{\textwidth}{X c}
\toprule
\textbf{Metric} & \textbf{Score} \\
\midrule
Precision & 0.773 \\
Recall & 0.731 \\
mAP@0.5 & 0.774 \\
mAP@0.5:0.95 & 0.295 \\
\bottomrule
\end{tabularx}
\caption{YOLOv8 Model Performance Metrics. The table summarizes key evaluation metrics including Precision, Recall, and mean Average Precision (mAP) across IoU thresholds.}
\label{tab:yolov8_metrics}
\end{table}

The annotated dataset used in this study is publicly available at: \href{https://github.com/karanS08/CGmap}{\texttt{https://github.com/karanS08/CGmap}}

\subsection{Detection, Centroid Extraction, and Point Cloud Generation}

A sliding window inference strategy was used to apply the trained YOLOv8n model to the full-resolution orthomosaics. Each detected bounding box was post-processed to compute its centroid:
\[
c_i = \left(x_i + \frac{w_i}{2},\; y_i + \frac{h_i}{2} \right)
\]
where $(x_i, y_i)$ and $(w_i, h_i)$ are the top-left corner and dimensions of the bounding box, respectively.

A {2D point cloud} was generated by projecting these centroids back into the global coordinate system (via georeferencing metadata):
\[
\mathcal{P} = \{c_1, c_2, \dots, c_n\}, \quad c_i \in \mathbb{R}^2
\]

Spatial mosaics were cropped through the reference of boundary polygons provided in Well-Known Text (WKT) format by the sugar mill, isolating individual farms. A 5-meter buffer was reinforced around each polygon for ensuring complete inclusion of edge rows, and prevention of detection losses due to spatial clipping.

The subsequent sections discuss the resulting point cloud datasets formed the basis for spatial analysis, including MST-based row extraction and bald spot detection.

\subsection{Spatial Normalization via Minimum Spanning Tree }

Let the set of detected plant centroids be:
\begin{equation}
\mathcal{P} = \{p_1, p_2, \dots, p_n\}, \quad p_i \in \mathbb{R}^2
\label{eq:points_set}
\end{equation}
where \(p_i = [x_i, y_i]^T\) denotes the 2D coordinates of the \(i^\text{th}\) plant.

\begin{figure}[H]
    \centering
    \includegraphics[width=\linewidth]{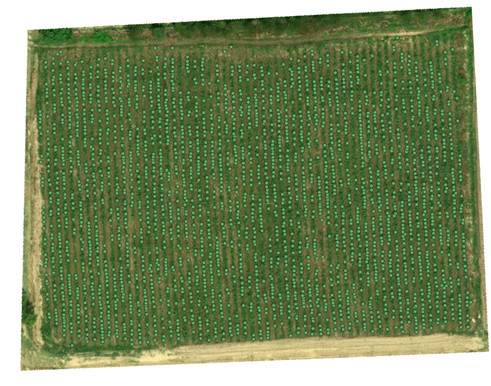}
    \caption{Image overlaid with AI detection for Sugarcane plant centroids, marked with green dots.}
    \label{fig:Centroids}
\end{figure}

Construct a fully connected undirected graph \(G = (\mathcal{P}, E)\), where each edge \(e_{ij} \in E\) connects nodes \(p_i\) and \(p_j\), and has a Euclidean distance weight:
\begin{equation}
w_{ij} = \| p_i - p_j \|_2
\label{eq:euclidean}
\end{equation}

Compute a Minimum Spanning Tree (MST) using a standard algorithm such as Prim's~\cite{MST} or Kruskal's~\cite{kruskal}, denoted:
\begin{equation}
T = (\mathcal{P}, E_T) \subseteq G
\label{eq:mst}
\end{equation}
such that:
\begin{equation}
\sum_{e_{ij} \in E_T} w_{ij} \quad \text{is minimized.}
\label{eq:mst_cost}
\end{equation}

To isolate intra-row connections (connections along the same planting row), apply a distance threshold:
\begin{equation}
w_{ij} < \theta, \quad \text{where } \theta = d_1 - \alpha_1
\label{eq:threshold}
\end{equation}
where:
\begin{itemize}
    \item \(d_1\) is the typical distance between adjacent rows (inter-row spacing),
    \item \(\alpha_1\) is a tolerance margin accounting for natural variability or measurement noise.
\end{itemize}

This filtering produces disconnected subgraphs \(T_1, T_2, \dots, T_k \subseteq T\), each corresponding to an individual planting row.

For each subgraph \(T_k\), define the unit direction vectors along edges:
\begin{equation}
\vec{u}_{ij} = \frac{p_j - p_i}{\| p_j - p_i \|_2}, \quad \forall e_{ij} \in T_k
\label{eq:unit_vector}
\end{equation}

The average row orientation vector is then:
\begin{equation}
\vec{v} = \frac{1}{|E_{T_k}|} \sum_{e_{ij} \in E_{T_k}} \vec{u}_{ij}
\label{eq:average_vector}
\end{equation}

The alignment angle with respect to the vertical axis is computed as:    
\begin{equation}
\theta^* = \arccos\left( \frac{\vec{v} \cdot \vec{y}}{\|\vec{v}\|} \right), \quad 
\vec{y} = \begin{bmatrix} 0 \\ 1 \end{bmatrix}
\label{eq:alignment_angle}
\end{equation}
where \(\vec{y}\) is the unit vector along the North–South direction.

\begin{figure}[H]
    \centering
    \includegraphics[width=\linewidth]{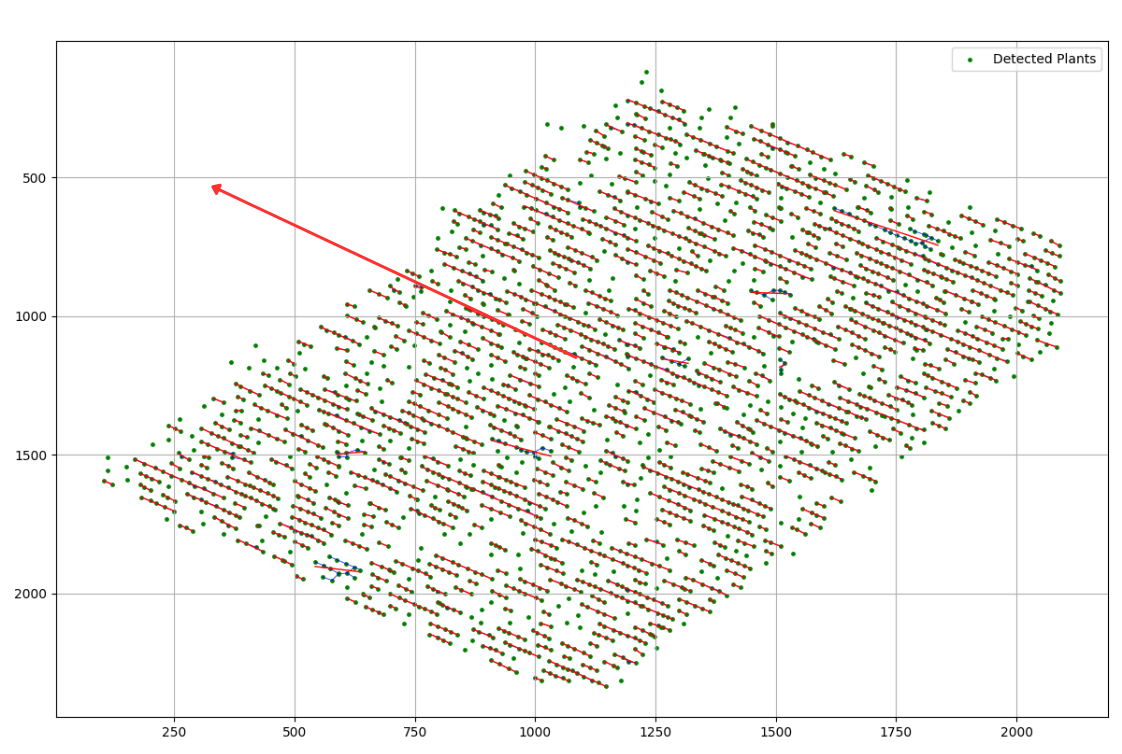}
    \caption{Point cloud marked with unit vector $\vec{y}$ along the North–South direction.}
    \label{fig:unit-vector}
\end{figure}

Finally, all points \(p_i\) are rotated by \(\theta^*\) (Eq.~\ref{eq:alignment_angle}) to yield a North–South aligned point cloud \(\mathcal{P}'\).

\begin{figure}[H]
    \centering
    \includegraphics[width=\linewidth]{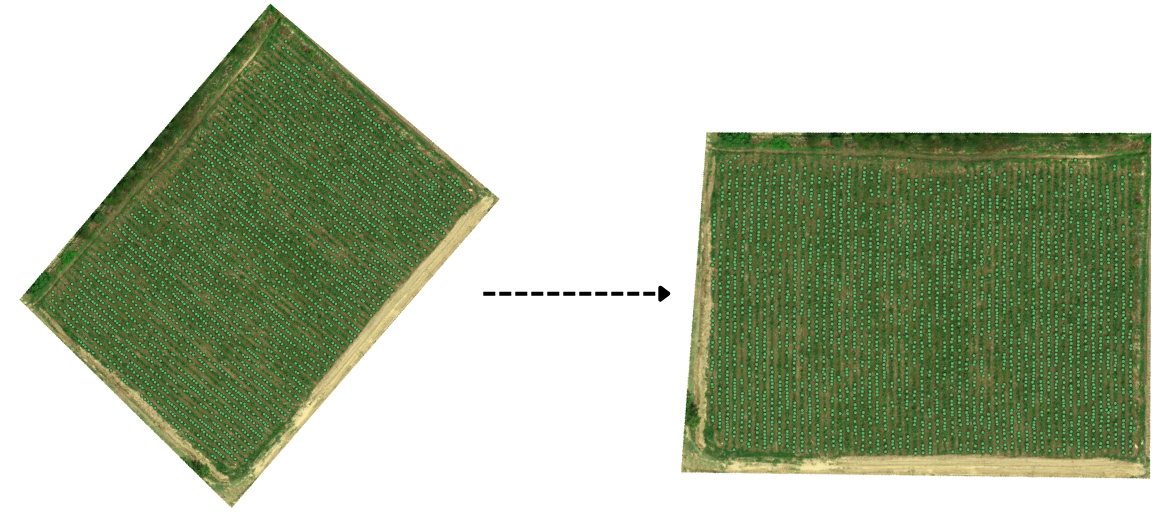}
    \caption{Spatially normalized image, \(\theta^* = 127.6^\circ\).}
    \label{fig:normalization}
\end{figure}

\subsection{Row and Column Extraction and Gap Detection}

After alignment, perform column detection on the rotated set \(\mathcal{P}'\) by applying a sweeping vertical-line algorithm to identify candidate column lines from the top-most and bottom-most points in each slice.

\begin{figure}[H]
    \centering
    \begin{subfigure}[b]{0.48\linewidth}
        \centering
        \includegraphics[width=\linewidth]{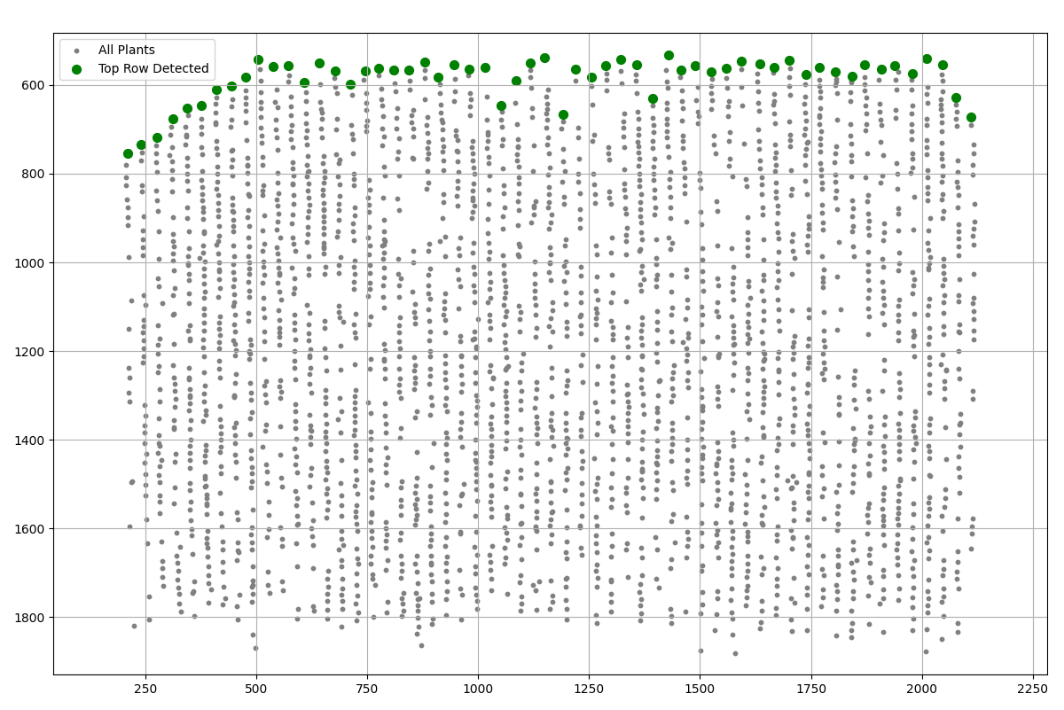}
        \caption{Sugarcane point cloud marked with topmost points for each slice.}
        \label{fig:sweeping-line}
    \end{subfigure}
    \hfill
    \begin{subfigure}[b]{0.48\linewidth}
        \centering
        \includegraphics[width=\linewidth]{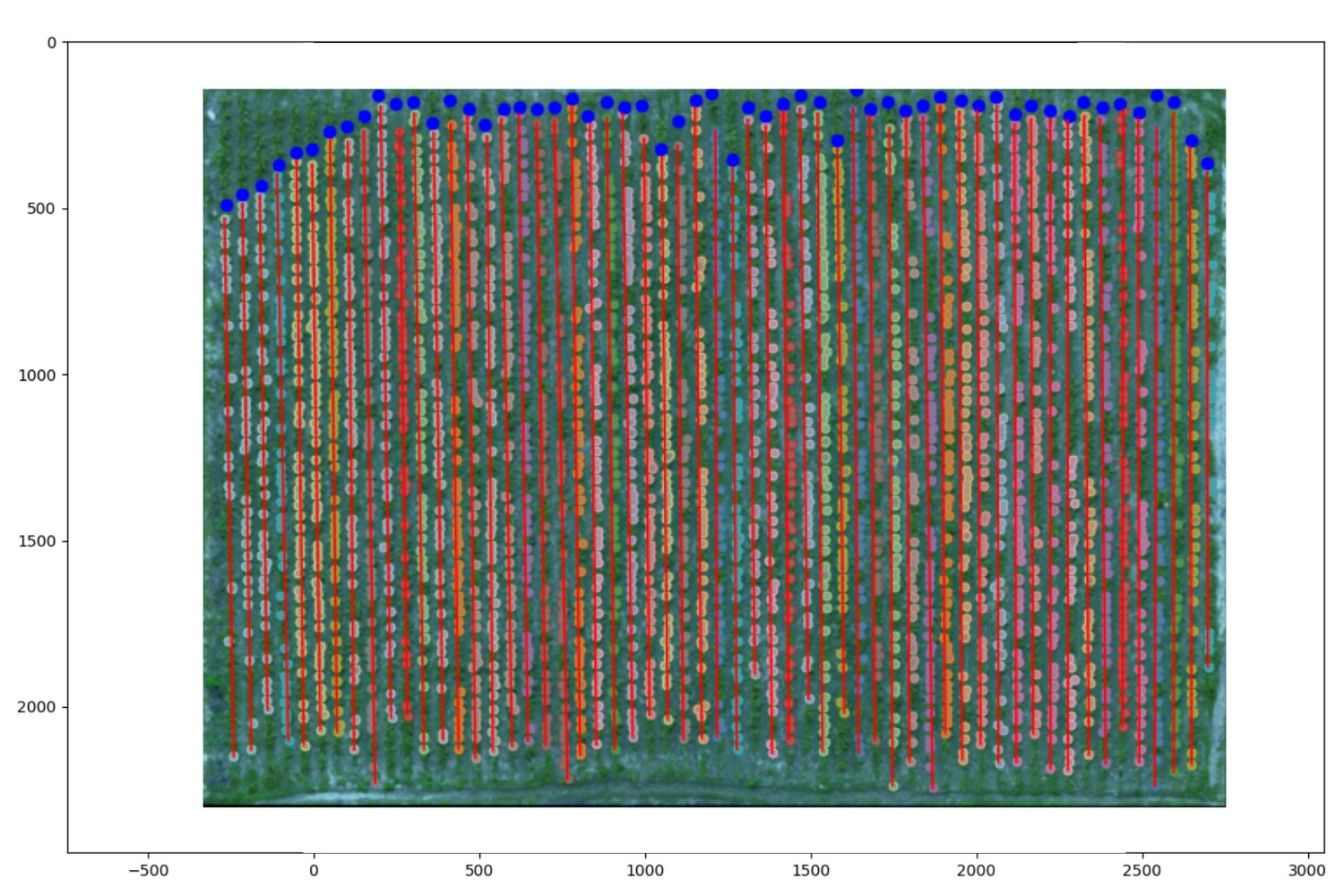}
        \caption{Column identification from sugarcane plant points using line fitting.}
        \label{fig:bestfit}
    \end{subfigure}
    \caption{(a) Line sweep for topmost points and (b) fitted lines for column identification in sugarcane plant point cloud.}
    \label{fig:combined}
\end{figure}

Each crop column is identified by first fitting a best-fit line to the plant points in that column using linear regression:
\begin{equation}
\hat{y} = mx + b
\label{eq:line_fit}
\end{equation}

A plant point is assigned to a column if its perpendicular distance to the fitted line is less than \(\frac{s_c}{2}\), where \(s_c\) is the expected column spacing.

For each column, the plant points are projected onto the fitted line, and the inter-plant distances are computed as:
\begin{equation}
d_i = \| p_{i+1}' - p_i' \|_2
\label{eq:inter_plant}
\end{equation}

Gaps are identified where:
\begin{equation}
d_i > d_2 + \alpha_2
\label{eq:gaps}
\end{equation}
where \(d_2\) is the expected plant-to-plant spacing and \(\alpha_2\) is a tolerance margin.

Each detected gap is stored as a geometric entity \(g_j \in \mathcal{G}\), forming the set:
\begin{equation}
\mathcal{G} = \{ g_1, g_2, \dots, g_m \}
\label{eq:gaps_set}
\end{equation}

The final spatial outputs consist of:
\begin{itemize}
    \item \(\mathcal{P}'\): detected plant centroids (North–South aligned),
    \item \(\mathcal{L}\): column lines,
    \item \(\mathcal{G}\): inferred gaps.
\end{itemize}

All spatial layers are exported in \texttt{WKT} format for GIS visualization and downstream agronomic analysis.

\begin{figure}[H]
    \centering
    \includegraphics[width=\linewidth]{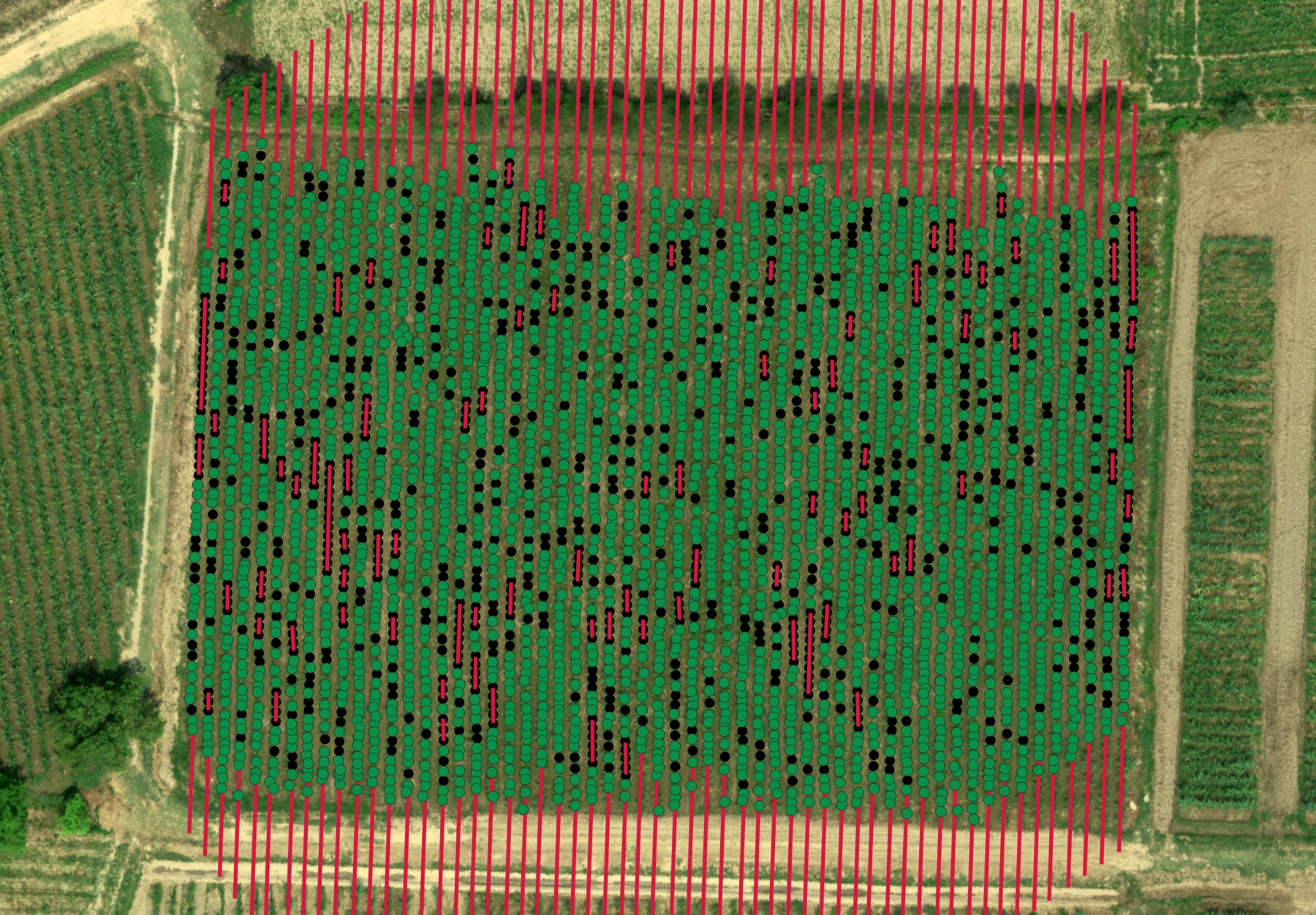}
    \caption{\texttt{WKT} file plotted on QGIS with the original orthomosaic as base map.}
    \label{fig:missing-map}
\end{figure}

\section{Results and discussion}

The proposed pipeline not only demonstrates strong robustness and adaptability across varied field layouts owing to its modular spatial analysis approach, but also can be inherently generalized as it relies on {point cloud abstraction} and {graph-based spatial reasoning}, unlike crop-specific heuristics. This system can also utilize MST clustering to extract local plant sequences and a global alignment strategy based on unit vector averaging, to reliably identify row and column structures in any crop with a grid-like planting pattern. 
	
The training dataset for the pipeline was obtained from multiple sugarcane farms in Western Uttar Pradesh, during the emergence phase of saplings (30–90 days after planting). The automated detection and gap-mapping workflow enabled rapid identification of bald spots across the surveyed fields, with field-scale germination maps generated within 30 minutes of drone flight processing.

Figure~\ref{fig:missing-map} depicts a representation of the output generated by the pipeline. While black dots represent one missing plant, indicating regions not requiring much intervention, red lines indicate major bald spots by highlighting series of consecutive missing plants (more than two plants), which might require more attention. This enables the identification of regions of high and low germination within the field, allowing field management backed by practical insights.

This system aims to facilitate timely transplantation interventions in underperforming zones by providing georeferenced maps of missing plants, promoting stand uniformity in corrected plots. In a yield assessment, substantial increase in average yield per acre is expected in fields where transplantation would be guided by the system's recommendations, in comparison to plots without early gap correction.

Moreover, the ability to integrate the output into existing GIS tools allows farm managers and sugar mills to coordinate harvesting and input supply logistics more efficiently. These findings highlight not only the agronomic value of the approach but also its {scalability and economic viability} for wide-scale adoption in commercial sugarcane production.

\section{Conclusion}
This work aims to contribute to the growing body of research on the use of object detection AI in agriculture through a scalable and interpretable vision based pipeline. In this particular use case it has been used for sugarcane detection in its early growth stages. It does so by employing UAV imagery with YOLOv8 based seedling detection and spacial reasoning on georeference point clouds. 

The overall performance of the pipeline has been successful in mapping bald spots on the field, aiding early detection which would facilitate early interventions to improve crop uniformity for better yields. Advantages offered by this approach are its modular and lightweight nature allowing for its implementation in resource constrained and edge computing environments (refer table \ref{tab:benchmark_summary}). Additionally, while unexplored, this pipeline is capable of naturally extending to crops following a planting geometry similar to sugarcane adding to its applicability. 

Future prospects of this research endeavour include- expansion of the dataset across regions, seasons and crop cycles for performance validation and improving the generalizability and robustness and benchmarking against ground truth emergence surveys.

\section*{Acknowledgments}
Our sincere gratitude to CNH industrial for their invaluable support in providing resources, funds and assistance for this essential research. Their involvement has been in instrumental to this collaboration. We would also like to thank Keshika Gajbhiye for proof reading the manuscript at various parts of preparation.
\bibliography{ref}  
\bibliographystyle{unsrt}

\appendix

\section{Edge deployment results}
\begin{table}[ht]
  \centering
  \begin{tabularx}{\linewidth}{l X r r}
    \toprule
    Machine & & FPS & Avg\_time (s) \\
    \midrule
    Intel i5-10500 & & 16.10 & 0.0621 \\
    GTX 1660 SUPER & & 69.82 & 0.0143 \\
    Raspberry Pi 4 (4GB) & & 0.94 & 1.0675 \\
    Raspberry Pi 5 (8GB) & & 2.09 & 0.7212 \\
    Intel Xeon W5 & & 19.63 & 0.0510 \\
    \bottomrule
  \end{tabularx}
  \caption{YOLOv8 inference performance across different hardware platforms, illustrating the model's scalability from high-performance GPUs to low-power edge devices.}
  \label{tab:benchmark_summary}
\end{table}

\end{document}